\documentclass[runningheads]{llncs}
\usepackage[T1]{fontenc}
\usepackage{graphicx,verbatim}
\usepackage{makecell}
\usepackage{multirow}
\usepackage{amsfonts}
\usepackage{amsmath}
\usepackage{mathrsfs}
\usepackage[bookmarks=true]{hyperref}
\usepackage{lipsum}
\usepackage{xcolor}
\usepackage{pgfplots}
\usepackage{enumitem}
\usepackage{subcaption}
\usepackage{colortbl}
\usepackage[misc]{ifsym}

\definecolor{gray}{RGB}{128,128,128}
\SetLipsumParListSurrounders{\begingroup\color{gray}}{\endgroup}

\definecolor{myred}{RGB}{255,0,0}
\definecolor{mygreen}{RGB}{0,176,80}

\usepackage{xspace}
\makeatletter
\DeclareRobustCommand\onedot{\futurelet\@let@token\@onedot}
\def\@onedot{\ifx\@let@token.\else.\null\fi\xspace}

\def\ie{\emph{i.e}\onedot}

\makeatother

\usepackage{color}

\begin{document}
\title{Cycle Context Verification for In-Context Medical Image Segmentation}
\titlerunning{Cycle Context Verification for In-Context Medical Image Segmentation}
%

\author{
    Shishuai Hu\inst{1,2}\thanks{S. Hu and Z. Liao contributed equally. Corresponding authors: H. Fu and Y. Xia.} \and 
    Zehui Liao\inst{1,2*} \and 
    Liangli Zhen\inst{2} \and 
    Huazhu Fu\inst{2}$^{(\textrm{\Letter})}$ \and 
    Yong Xia\inst{1,3,4}$^{(\textrm{\Letter})}$
} 
\authorrunning{S. Hu et al.}
\institute{
    National Engineering Laboratory for Integrated Aero-Space-Ground-Ocean Big Data Application Technology, School of Computer Science and Engineering, Northwestern Polytechnical University, Xi’an 710072, China \\
    \and
    Institute of High Performance Computing, Agency for Science, Technology and Research, Singapore 138632, Singapore \\
    \and
    Ningbo Institute of Northwestern Polytechnical University, Ningbo 315048, China\\
    \and
    Research \& Development Institute of Northwestern Polytechnical University in Shenzhen, Shenzhen 518057, China\\
    \email{hzfu@ieee.org; yxia@nwpu.edu.cn} \\
}

\maketitle              
\begin{abstract}
In-context learning (ICL) is emerging as a promising technique for achieving universal medical image segmentation, where a variety of objects of interest across imaging modalities can be segmented using a single model.
Nevertheless, its performance is highly sensitive to the alignment between the query image and in-context image-mask pairs.
In a clinical scenario, the scarcity of annotated medical images makes it challenging to select optimal in-context pairs, and fine-tuning foundation ICL models on contextual data is infeasible due to computational costs and the risk of catastrophic forgetting.
To address this challenge, we propose Cycle Context Verification (CCV), a novel framework that enhances ICL-based medical image segmentation by enabling self-verification of predictions and accordingly enhancing contextual alignment. 
Specifically, CCV employs a cyclic pipeline in which the model initially generates a segmentation mask for the query image. Subsequently, the roles of the query and an in-context pair are swapped, allowing the model to validate its prediction by predicting the mask of the original in-context image. 
The accuracy of this secondary prediction serves as an implicit measure of the initial query segmentation. 
A query-specific prompt is introduced to alter the query image and updated to improve the measure, thereby enhancing the alignment between the query and in-context pairs. 
We evaluated CCV on seven medical image segmentation datasets using two ICL foundation models, demonstrating its superiority over existing methods. 
Our results highlight CCV’s ability to enhance ICL-based segmentation, making it a robust solution for universal medical image segmentation.
The code will be available at \url{https://github.com/ShishuaiHu/CCV}.


\end{abstract}
\section{Introduction}
Universal medical image segmentation has garnered significant attention due to the increasing need for medical professionals to segment and analyze various objects of interest across diverse imaging modalities, rather than focusing on a single organ or lesion~\cite{ma2024segment,butoi2023universeg,zhao2024foundation,kirillov2023segment,mazurowski2023segment,huang2024segment,chen2024ma,mei2025survey,zeng2024human}. 
In-context learning (ICL)~\cite{dong2024survey}, which enables a single model to segment varied targets through contextual guidance\cite{butoi2023universeg}, has shown promise in addressing this challenge.

Initially developed for large language models (LLMs), ICL leverages contextual information to improve predictions~\cite{dong2024survey} and has been successfully adapted to both natural~\cite{bar2022visual,wang2023seggpt,wang2023images} and medical image segmentation~\cite{butoi2023universeg}.
In a typical ICL framework for image segmentation, a test image (query) is processed using in-context image-mask pairs that exemplify the target structure.
When trained on diverse, large-scale datasets, these models exhibit universal segmentation capabilities~\cite{bar2022visual,wang2023seggpt,wang2023images,butoi2023universeg}.
Despite their potential, the performance of these models is highly sensitive to the alignment between the query image and the in-context pairs~\cite{zhang2023makes}. Consequently, many research efforts have been devoted to 
selecting optimal in-context pairs for each test query ~\cite{xu2024towards,gao2024boosting,wu2024efficient,suo2024rethinking,sun2023exploring}.
However, the ideal context for a given query may not exist within the available in-context pairs. Moreover, in medical image segmentation, particularly when dealing with underrepresented targets, the availability of suitable in-context pairs is often limited~\cite{cheng2024few}, making their selection impractical. Furthermore, fine-tuning the ICL model on these in-context pairs is not feasible due to computational costs and the risk of catastrophic forgetting~\cite{lin2023speciality}.
In this paper, we suggest an alternative approach: enhancing the context rather than selecting optimal pairs or fine-tuning the ICL model to improve segmentation performance.

Current methods for visual context enhancement in ICL are limited. One relevant approach is InMeMo~\cite{zhang2024instruct}, which introduces task-specific prompts added to the borders of the context image to guide the ICL model during segmentation. However, InMeMo requires a substantial amount of labeled data to train these task-specific prompts, and the data must correspond to the same task as the test query image. This is particularly impractical in medical image segmentation due to the scarcity of suitable in-context pairs~\cite{cheng2024few}.
To overcome this limitation, we propose learning a query-specific prompt to enhance the context for each query image and optimize it with the limited available in-context pairs during testing. However, defining an optimization objective at test time to fully exploit labeled in-context pairs remains a challenge.

\begin{figure}[tbp]
    \centering
    \includegraphics[scale=0.6]{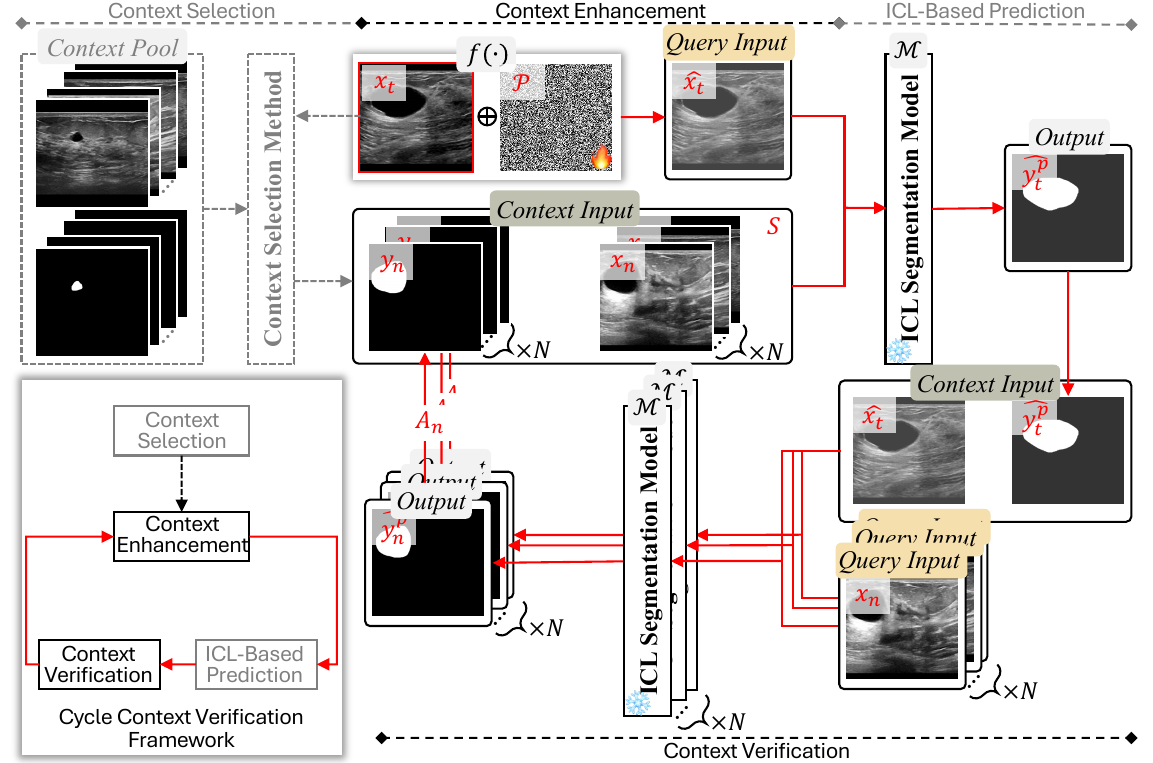}
    \caption{
    Diagram of the proposed CCV framework. The symbol $\oplus$ denotes the addition of the query-specific prompt to the query image. The red arrows represent the cycle data flow.
    }
    \label{fig:overview}
\end{figure}

In the field of LLMs, test time scaling is gaining increasing attention~\cite{snell2024scaling,muennighoff2025s1simpletesttimescaling}.
This approach incorporates constraints, such as budget forcing~\cite{muennighoff2025s1simpletesttimescaling}, to prompt the model to verify its outputs and correct reasoning errors.
Inspired by the concept of answer verification, we propose a cycle context verification mechanism (see ~\figurename{~\ref{fig:overview}}), which allows the ICL model to double-check its predicted mask and optimize the query-specific prompt to improve segmentation.
Specifically, given a test query image and an associated in-context pair, the ICL model generates an initial segmentation mask for the query.
We then swap the roles of the query and in-context pair: the query image and its predicted mask become the new context pair, while the original in-context image becomes the new query image.
By taking the swapped query and context pair as input, the ICL model can double-check its original prediction for the query image and generate a mask prediction for the in-context image. 
The accuracy of the original query prediction indicates the alignment between the query image and the context pair, which also impacts the accuracy of the in-context image prediction.
In other words, the accuracy of the in-context prediction implies the accuracy of the original query prediction.
Since the in-context mask is available, the accuracy of the in-context prediction can be directly evaluated.
This enables optimization of the query-specific prompt, thus improving the accuracy of the in-context prediction and, in turn, enhancing the query segmentation.

In this paper, we introduce Cycle Context Verification (CCV) for in-context medical image segmentation.
CCV employs a cyclic pipeline during testing that allows the ICL model to verify its query prediction. A learnable, query-specific prompt is incorporated into the query image to improve the alignment between the query and the in-context pairs. This prompt is iteratively updated to improve the accuracy of the in-context prediction, leading to better query segmentation. We evaluate the proposed method against other techniques using two ICL backbones on seven medical image segmentation datasets, demonstrating its superior performance over existing methods.
The contributions of this work are three-fold: 
\begin{enumerate}
    \item We propose a novel, plug-and-play cycle context verification pipeline that enables the ICL model to double-check its query predictions at inference time.
    \item Unlike traditional prompt learning, which tunes a unified prompt across the entire dataset, we introduce a learnable, query-specific prompt that is optimized to improve the alignment between each query image and its in-context pairs, thereby enhancing segmentation performance.
    \item Our experiments on seven medical image segmentation datasets, using two ICL foundation models, show that the proposed method outperforms existing techniques.
\end{enumerate}

\section{Method}
\subsection{Problem Formulation and Method Overview}
Let $\mathcal{M}$ denote a pretrained ICL model for medical image segmentation, and let $S = \{(x_n, y_n)\}_{n=1}^N$ represent a set of in-context image-mask pairs, where $N$ is the number of available pairs. 
$S$ can be the limited available pairs or context pairs selected through popular context selection methods~\cite{zhang2023makes,radford2021learning,gao2024boosting}.
Given a test query image $x_t$, the objective is to enhance the alignment between $S$ and $x_t$ to improve the predicted segmentation mask $\hat{y_t^p} = \mathcal{M}(f(x_t), S)$, while keeping model $\mathcal{M}$ frozen. Here, $f(\cdot)$ represents a function that modifies $x_t$.

The proposed CCV framework introduces a Cycle ICL pipeline, which enables the model to double-check its query prediction $y_t^p$ iteratively, a proxy accuracy measurement $\mathcal{A}$ to evaluate the accuracy of $y_t^p$, and a learnable query-specific prompt $\mathcal{P}$ to refine $x_t$ towards generating more accurate $y_t^p$. 
In the Cycle ICL pipeline, a segmentation mask $y_t^p$ is firstly predicted by $\mathcal{M}$ as usual. 
The model then swaps the roles of the query and in-context pair, treating the query and its predicted mask as a new in-context pair while designating the original in-context image as the new query. This process yields a secondary prediction $y_n^p$, whose accuracy, measured using $\mathcal{A}$, serves as an indirect estimate of the reliability of $y_t^p$.
The query-specific prompt $\mathcal{P}$ is iteratively optimized to maximize the accuracy measurement $\mathcal{A}$. Once converged, the enhanced inputs $f(x_t)$ and $S$ are fed to $\mathcal{M}$ to yield the final prediction $\hat{y_t^p}$.
An overview of the framework is illustrated in~\figurename{~\ref{fig:overview}}. We now delve into its details.

\subsection{Cycle ICL Pipeline for Context Verification}
Given a set of in-context pairs $S = \{(x_n, y_n)\}_{n=1}^N$, a pretrained ICL model $\mathcal{M}$, and a query image $x_t$, the initial query prediction $y_t^p$ is generated as:
\begin{equation}
  y_t^p = \mathcal{M}(x_t, S).
\label{eq:ytp}
\end{equation}

To enable self-verification, the model's input configuration is modified. The predicted mask $y_t^p$ is paired with the original query image $x_t$ to create a new in-context pair $\{(x_t, y_t^p)\}$, while the original in-context image $x_n$ is designated as the new query. The model then generates a secondary prediction $y_n^p$:
\begin{equation}
  y_n^p = \mathcal{M}(x_n, \{(x_t, y_t^p)\}).
  \label{eq:ynp}
\end{equation}

Since the ground truth mask $y_n$ for $x_n$ is available, the accuracy of $y_n^p$ can be quantitatively evaluated using the Dice Similarity Coefficient (DSC) $\mathcal{A}_n=\mathcal{D}(y_n^p, y_n)$.
The overall accuracy $\mathcal{A}$ across all in-context pairs is then computed as:
\begin{equation}
 \mathcal{A} = \frac{\sum_{n=1}^N \mathcal{A}_n}{N}.
 \label{eq:a}
\end{equation}

The segmentation quality of $\mathcal{M}$ is contingent on the context provided. If the initial query prediction $y_t^p$ deviates substantially from the ground truth, incorporating it into the context may degrade the accuracy of $y_n^p$. Thus, $\mathcal{A}$ serves as an implicit measure of the reliability of $y_t^p$.

\subsection{Query-Specific Prompt for Context Enhancement}
To improve segmentation accuracy, the objective is to maximize $\mathcal{A}$. However, directly optimizing $\mathcal{M}$ may potentially lead to model collapse, where the model becomes overfitted to the verification step without improving $y_t^p$.
To circumvent this, we introduce a query-specific prompt $\mathcal{P}$, which is learnable and spatially aligned with the query image $x_t$. By adding $\mathcal{P}$ to $x_t$, the correspondence between the query and in-context images is expected to be enhanced, thereby yielding improved segmentation outcomes. The transformation of the query image can be expressed as:
\begin{equation}
 \hat{x_t} = f(x_t) = x_t + \mathcal{P}.
\end{equation}

Using the modified query $\hat{x_t}$, we recompute $\hat{y_t^p}$, $\hat{y_n^p}$, $\hat{\mathcal{A}_n}$, and $\hat{\mathcal{A}}$. The loss function is defined as:
\begin{equation}
 \mathcal{L} = 1 - \hat{\mathcal{A}}.
\end{equation}

Minimizing $\mathcal{L}$ encourages the model to refine $\mathcal{P}$ iteratively. A higher $\hat{\mathcal{A}}$ indicates an improved performance of the predicted masks $\{\hat{y_n^p}\}_{n=1}^N$, thereby improving the reliability of ${y_t^p}$. To maintain computational efficiency, an accuracy threshold $\mathcal{T}$ is established. If $\hat{\mathcal{A}} > \mathcal{T}$, prompt optimization is halted. Additionally, a maximum iteration limit $\mathcal{E}$ is imposed to prevent excessive optimization. 
Once $\mathcal{P}$ is optimized, the final segmentation mask $\hat{y_t^p}$ is generated by feeding $\hat{x_t}$ and $S$ into $\mathcal{M}$.

\section{Experiments and Analysis}
\subsection{Experimental Setup}
\noindent
\textbf{ICL Backbones.}
The proposed method is evaluated using UniverSeg~\cite{butoi2023universeg} and SegGPT~\cite{wang2023seggpt} as ICL backbones. 
UniverSeg is an ICL backbone specifically designed for medical image segmentation. 
SegGPT, on the other hand, is an ICL backbone primarily trained on large-scale natural image segmentation datasets. 

\noindent
\textbf{Datasets and Evaluation Metrics.}
We selected seven medical image segmentation datasets encompassing eight tasks across seven medical imaging modalities\footnote{These datasets can be accessed through \url{https://huggingface.co/datasets/microsoft/BiomedParseData}.}~\cite{zhao2024foundation} to evaluate the proposed method. 
These include the 
BreastUS (BUS, 647 images)~\cite{al2020dataset}, 
COVID-QU-Ex (CQE, 5826 images)~\cite{tahir2021covid}, 
GlaS (GLS, 165 images)~\cite{sirinukunwattana2015stochastic}, 
NeoPolyp (NPP, 1000 images)~\cite{ngoc2021neounet}, 
OCT-CME (OCT, 1460 images)~\cite{ahmed2022deep}, 
REFUGE (ROC and ROD, 1200 images)~\cite{orlando2020refuge}, and 
UWaterlooSkinCancer (USC, 206 images)~\cite{venugopal2022dtp} 
datasets. 
Each dataset is partitioned into proportions of 3:1:1, designated for test samples, validation samples, and candidate in-context samples, respectively.
Notably, these datasets are not included in the training sets of either UniverSeg or SegGPT. 
The DSC is used as the evaluation metric to assess the segmentation performance.

\noindent
\textbf{Implementation Details.}
The proposed method is implemented using the PyTorch framework and tested on a single NVIDIA 3090 GPU.
For the UniverSeg model, the context size is set to eight. All input images are resized to 128 $\times$ 128, converted to gray-scale images, and normalized to the intensity range of [0, 1].
For the SegGPT backbone, the context size is set to one. All images are resized to 448 $\times$ 448, converted to three-channel color images, and normalized by subtracting the mean and dividing by the standard deviation.
The query-specific prompt $\mathcal{P}$ is zero-initialized.
The threshold $\mathcal{T}$ and the iteration limitation $\mathcal{E}$ are empirically set to 0.9 and 20 respectively.

\begin{table}[tbp]
  \setlength\tabcolsep{2pt}
  \centering
  \caption{
    The DSC (\%) of the competing methods and those equipped with CCV. 
  }
  \label{tab:comparision}
  \begin{tabular}{c|c|c|c|c|c|c|c|c|c|c}
    \hline
    \hline
    $\mathcal{M}$  & Methods   & BUS   & CQE   & GLS   & NPP   & OCT   & ROC   & ROD   & USC    & Average \\ 
    \hline
    \hline
    \multirow{8}{*}{\rotatebox{90}{UniverSeg~\cite{butoi2023universeg}}}  
    & RS         & 47.49 & 87.21 & 60.50 & 28.84 & 34.97 & 57.72 & 77.83 & 80.79 & 59.42 \\ \cline{2-11}
    & \cellcolor{green!10}RS+CCV    & \cellcolor{green!10}50.99 & \cellcolor{green!10}87.79 & \cellcolor{green!10}60.71 & \cellcolor{green!10}33.52 & \cellcolor{green!10}38.63 & \cellcolor{green!10}60.26 & \cellcolor{green!10}79.65 & \cellcolor{green!10}81.92 & \cellcolor{green!10}$61.68_{\textcolor{red}{\uparrow2.26}}$ \\ \cline{2-11} 
    & VPR~\cite{zhang2023makes}        & 54.13 & 90.68 & 60.06 & 32.34 & 44.72 & 61.25 & 78.76 & 83.91 & 63.23 \\ \cline{2-11} 
    & \cellcolor{green!10}VPR+CCV   & \cellcolor{green!10}57.28 & \cellcolor{green!10}91.11 & \cellcolor{green!10}62.82 & \cellcolor{green!10}36.38 & \cellcolor{green!10}47.73 & \cellcolor{green!10}63.39 & \cellcolor{green!10}80.27 & \cellcolor{green!10}85.22 & \cellcolor{green!10}$65.53_{\textcolor{red}{\uparrow2.30}}$ \\ \cline{2-11} 
    & DualSC~\cite{gao2024boosting}        & 53.64 & 88.60 & 62.29 & 34.47 & 46.34 & 62.10 & 79.30 & 84.07 & 63.85 \\ \cline{2-11} 
    & \cellcolor{green!10}DualSC+CCV   & \cellcolor{green!10}57.58 & \cellcolor{green!10}91.11 & \cellcolor{green!10}62.91 & \cellcolor{green!10}36.84 & \cellcolor{green!10}48.03 & \cellcolor{green!10}63.66 & \cellcolor{green!10}80.44 & \cellcolor{green!10}85.11 & \cellcolor{green!10}$65.71_{\textcolor{red}{\uparrow 1.86}}$ \\ \cline{2-11} 
    & InMeMo~\cite{zhang2024instruct}        & 48.01 & 86.73 & 61.98 & 28.56 & 34.53 & 54.41 & 79.33 & 81.27 & 59.35 \\ \cline{2-11} 
    & \cellcolor{green!10}InMeMo+CCV   & \cellcolor{green!10}51.67 & \cellcolor{green!10}87.34 & \cellcolor{green!10}62.15 & \cellcolor{green!10}31.53 & \cellcolor{green!10}37.34 & \cellcolor{green!10}58.01 & \cellcolor{green!10}80.11 & \cellcolor{green!10}82.37 & \cellcolor{green!10}$61.32_{\textcolor{red}{\uparrow1.97}}$ \\ 
    \hline
    \hline
    \multirow{8}{*}{\rotatebox{90}{SegGPT~\cite{wang2023seggpt}}}  
    & RS         & 42.05 & 91.00 & 76.66 & 59.89 & 35.85 & 67.28 & 91.38 & 80.88 & 68.12 \\ \cline{2-11} 
    & \cellcolor{green!10}RS+CCV    & \cellcolor{green!10}46.26 & \cellcolor{green!10}91.79 & \cellcolor{green!10}80.76 & \cellcolor{green!10}64.96 & \cellcolor{green!10}37.67 & \cellcolor{green!10}71.12 & \cellcolor{green!10}92.46 & \cellcolor{green!10}84.13 & \cellcolor{green!10}$71.14_{\textcolor{red}{\uparrow3.02}}$ \\ \cline{2-11} 
    & VPR~\cite{zhang2023makes}       & 55.92 & 93.44 & 85.09 & 68.29 & 58.26 & 68.06 & 91.91 & 82.64 & 75.45 \\ \cline{2-11} 
    & \cellcolor{green!10}VPR+CCV   & \cellcolor{green!10}60.71 & \cellcolor{green!10}93.85 & \cellcolor{green!10}86.46 & \cellcolor{green!10}72.04 & \cellcolor{green!10}59.32 & \cellcolor{green!10}72.13 & \cellcolor{green!10}92.59 & \cellcolor{green!10}85.54 & \cellcolor{green!10}$77.83_{\textcolor{red}{\uparrow2.38}}$ \\ \cline{2-11} 
    & DualSC~\cite{gao2024boosting}        & 57.43 & 93.55 & 85.94 & 68.97 & 53.54 & 70.92 & 92.00 & 83.23 & 75.70 \\ \cline{2-11} 
    & \cellcolor{green!10}DualSC+CCV   & \cellcolor{green!10}60.79 & \cellcolor{green!10}93.88 & \cellcolor{green!10}86.73 & \cellcolor{green!10}71.58 & \cellcolor{green!10}60.64 & \cellcolor{green!10}72.19 & \cellcolor{green!10}92.62 & \cellcolor{green!10}85.38 & \cellcolor{green!10}$77.98_{\textcolor{red}{\uparrow 2.28}}$ \\ \cline{2-11} 
    & InMeMo~\cite{zhang2024instruct}        & 59.68 & 95.37 & 81.85 & 63.50 & 46.42 & 78.75 & 93.22 & 82.83 & 75.20 \\ \cline{2-11} 
    & \cellcolor{green!10}InMeMo+CCV   & \cellcolor{green!10}61.77 & \cellcolor{green!10}95.57 & \cellcolor{green!10}83.09 & \cellcolor{green!10}66.26 & \cellcolor{green!10}46.85 & \cellcolor{green!10}82.18 & \cellcolor{green!10}94.03 & \cellcolor{green!10}86.20 & \cellcolor{green!10}$76.99_{\textcolor{red}{\uparrow1.79}}$ \\ 
    \hline
    \hline
\end{tabular}
  \end{table}

\subsection{Experimental Results and Analysis}
We employ three in-context pair selection methods as baselines to select in-context pairs from the candidate in-context dataset for each test query sample, including: Random Selection (RS), which involves the random selection of in-context pairs from the candidate in-context dataset, 
VPR~\cite{zhang2023makes}, which fine-tunes a pretrained CLIP image encoder~\cite{radford2021learning} to extract and rank features, subsequently selecting the top-K in-context pairs,
and DualSC~\cite{gao2024boosting}, which incorporates mask similarity into the ranking process upon VPR.
We also incorporate InMeMo~\cite{zhang2024instruct}, a context enhancement method that prompts the model regarding the specific segmentation task to be performed, as a baseline method. 
The proposed method is designed to be complementary to these existing approaches. Therefore, we integrate the proposed method with these baselines and report the 
performance. 

\noindent
\textbf{Comparison using UniverSeg as Backbone.}
The Dice scores attained by other competing methods, as well as those equipped with the proposed method, are presented in the first part of~\tablename{~\ref{tab:comparision}}. 
The results indicate that our method 
improves the overall segmentation performance of RS, VPR, and DualSC, emphasizing the necessity of enhancing context during inference.
Furthermore, RS equipped with our method outperforms InMeMo, affirming that query-specific prompts can be superior to task-specific prompts.
Our approach also enhances the performance of InMeMo, indicating that the query-specific prompt design is complementary to task-specific prompt.

\noindent
\textbf{Comparison using SegGPT as Backbone.}
The Dice scores for our model and competing methods using SegGPT as backbone are presented in the second part of ~\tablename{~\ref{tab:comparision}}. Although using fewer in-context pairs, SegGPT still outperforms UniverSeg, highlighting the strong generalization capabilities of SegGPT. 
Even though, the proposed method enhances the overall segmentation performance across all competing approaches, further demonstrating its advantages.

\noindent
\begin{minipage}{\textwidth}
\begin{minipage}[t]{0.43\textwidth}
\makeatletter\def\@captype{table}
\setlength{\belowcaptionskip}{7pt}
\renewcommand\arraystretch{0.9}
\centering
\setlength\tabcolsep{3pt}
\caption{
Performance of removing cycle ICL pipeline and prompt.
}
\label{tab:ablation}
\begin{tabular}{c|c|c}
\hline
\hline
\multicolumn{2}{c|}{Method} & \multirow{2}{*}{Average} \\ \cline{1-2}
\makecell{Cycle ICL\\Pipeline} & \makecell{Prompt\\Optimiz.} &   \\ \hline
\hline
$\times$ & $\times$ & 58.34 \\ \hline
\checkmark & $\times$ & 57.24 \\ \hline
$\times$ & \checkmark & 58.57 \\ \hline
\checkmark & \checkmark & 60.65 \\ \hline
\hline
\end{tabular}
\end{minipage}
\begin{minipage}[t]{0.54\textwidth}
\makeatletter\def\@captype{table}
\setlength{\belowcaptionskip}{7pt}
\renewcommand\arraystretch{0.9}
\centering
\setlength\tabcolsep{2pt}
\caption{
Performance of four variants of prompt optimization.
}
\label{tab:prompt}
\begin{tabular}{c|c|c|c|c}
\hline
\hline
\multicolumn{4}{c|}{Method} & \multirow{2}{*}{Average} \\ \cline{1-4}
\makecell{Border\\Prompt} & \makecell{Image\\Prompt} & \makecell{Update\\First} & \makecell{Update\\Second} &  \\ \hline
\hline
$\times$ & \checkmark & \checkmark & $\times$ & 59.76 \\ \hline
$\times$ & \checkmark & $\times$ & \checkmark & 59.37 \\ \hline
$\times$ & \checkmark & \checkmark & \checkmark & 60.65 \\ \hline
\checkmark & $\times$ & \checkmark & \checkmark & 58.82 \\ \hline
\hline
\end{tabular}
\end{minipage}
\end{minipage}

\subsection{Ablation Analysis}
To evaluate the effectiveness of the proposed method in a limited context scenario, we select a small number of in-context pairs, \ie, eight for UniverSeg, from the candidate in-context dataset for each task, utilizing them as context to guide the segmentation and evaluate the performance of the ICL model and the proposed method.
The effectiveness of the cycle ICL pipeline and the query-specific prompt optimization is also evaluated in the limited context scenario via removing these two components respectively.
UniverSeg is used as the ICL backbone for the ablation experiments.
The results are shown in~\tablename{~\ref{tab:ablation}}.
Removing prompt optimization represents employing solely the cycle ICL pipeline and updating the entire ICL model with a small learning rate 1e-4 for 20 iterations.
Removing Cycle ICL pipeline represents employing only the query-specific prompt and minimizing the entropy of $\hat{y_t^p}$ to update $\mathcal{P}$.
The performance drop of removing prompt optimization can be attributed to the model collapse when updating the entire ICL model. 
The results reveal that both the cyclical design and the query-specific prompt-based optimization are essential for the functioning of the proposed method.

\begin{figure}[tbp]
    \centering
    \includegraphics[scale=0.29]{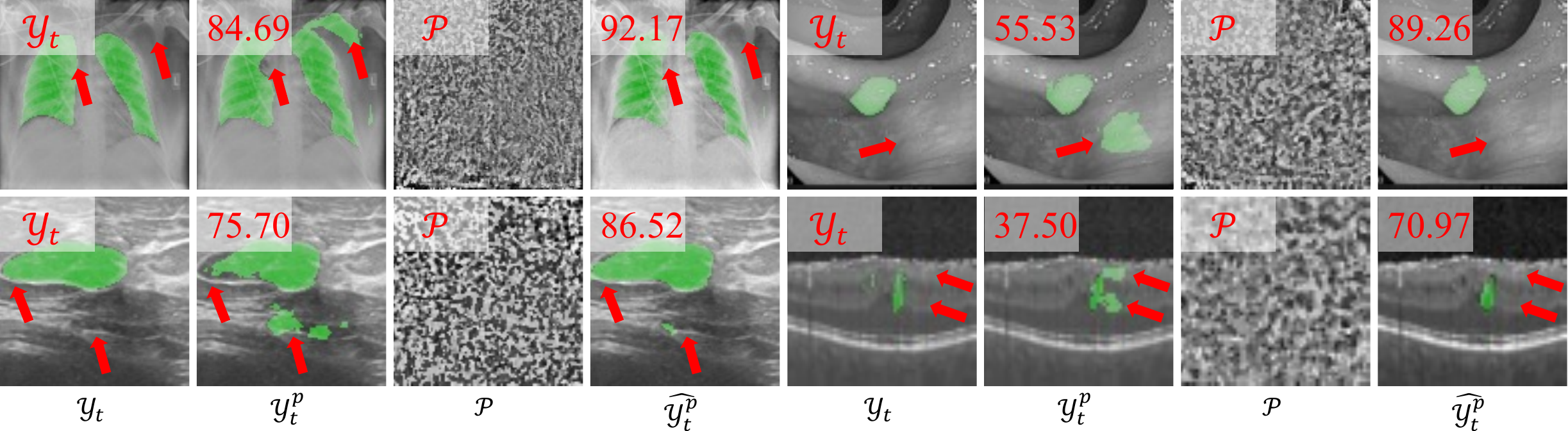}
    \caption{
    Visualization of four learned query-specific prompt $\mathcal{P}$, corresponding predictions $\hat{y_t^p}$ and original predictions $y_t^p$. Together with ground truth $y_t$. The dice scores of $y_t^p$ and $\hat{y_t^p}$ is also displayed for comparison. Note that $\mathcal{P}$ has been rescaled to [0, 255] for a better visualization. Best viewed in color.
    }
    \label{fig:prompt_vis}
\end{figure}

\noindent
\textbf{Analysis of Query-specific Prompt.}
We compare four configurations in the limited context scenario using the cycle ICL pipeline: use image prompt (CCV), use border prompt~\cite{zhang2024instruct} instead of image prompt, update the image prompt only in the first stage when generating $\hat{y_t^p}$,
and update the image prompt only in the second stage when generating $\hat{y_n^p}$.
The results is shown in~\tablename{~\ref{tab:prompt}}.
It indicates that the performance of using border prompt is inferior to that of using image prompt.
Notably, freezing the image prompt during the generation of either $\hat{y_t^p}$ or $\hat{y_n^p}$ is found to be less effective than updating the image prompt in these two stages.
This observation underscores the necessity for the updating of $\mathcal{P}$ to enhance the alignment between the query image and in-context pairs throughout these stages.
We visualized examples of the learned query-specific prompt $\mathcal{P}$ in~\figurename{~\ref{fig:prompt_vis}}. 
The predictions generated by the ICL model for $x_t$ and $\hat{x_t}$ are also illustrated.
Additionally, the ground truth segmentation maps are provided for reference.
The visualization demonstrates that the learned $\mathcal{P}$ can effectively alter $x_t$ towards generating markedly improved segmentation results, thereby underscoring the validity of the proposed methodology.

\section{Conclusion}
In this paper, we introduced the Cycle Context Verification (CCV) framework to enhance in-context medical image segmentation, particularly under limited in-context data. 
By implementing a cycle ICL pipeline, the ICL model is capable of double-checking its predictions.
The integration and optimization of a query-specific prompt enhance the alignment between the test query image and the associated in-context pairs.
Experimental results demonstrate that CCV not only improves existing context selection and enhancement methods, but also elevates segmentation performance in challenging limited context scenarios. 
Furthermore, ablation studies elucidate the design of the cycle ICL pipeline and query-specific prompt-based optimization.
These findings highlight the potential of CCV to advance in-context learning based universal medical image segmentation.

\begin{credits}
\subsubsection*{\ackname} This work was supported 
in part by the National Natural Science Foundation of China under Grants 62171377 and 92470101, 
in part by the ``Pioneer'' and ``Leading Goose'' R\&D Program of Zhejiang, China, under Grant 2025C01201(SD2), 
in part by the Ningbo Clinical Research Center for Medical Imaging under Grant 2021L003 (Open Project 2022LYKFZD06), 
in part by the Shenzhen Science and Technology Program under Grant JCYJ20220530161616036, and 
in part by the Innovation Foundation for Doctor Dissertation of Northwestern Polytechnical University under Grants CX2023016 and CX2022056.
\end{credits}

%
%
%
\bibliographystyle{splncs04}
\bibliography{reference.bib}
\end{document}